%% file: main.tex
\documentclass[10pt]{article} 
\usepackage[preprint]{tmlr}
\input{math_commands.tex}

\usepackage{graphicx}

\usepackage{amssymb}
\usepackage{pifont}
\newcommand{\cmark}{\ding{51}}%
\newcommand{\xmark}{\ding{55}}%
\usepackage{multirow, booktabs}
\usepackage{makecell}

\usepackage{hyperref}
\usepackage{url}

\title{Fast Image-based Neural Relighting with Translucency-Reflection Modeling\thanks{
This manuscript is an updated version of our preprint arXiv:2306.09322v1.
The earlier version (v1) contained an implementation error in the evaluation of the NRTF baseline,
leading to incorrect reported results.
In this version, we correct that error, and all experimental results in this paper supersede those reported in v1.
}}

\author{\name Shizhan Zhu, \name Shunsuke Saito, \name Aljaz Bozic, \name Carlos Aliaga, \name Trevor Darrell, \name Christoph Lassner}

\def\month{MM}  
\def\year{YYYY} 
\def\openreview{\url{https://openreview.net/forum?id=XXXX}} 

\begin{document}

\maketitle

\begin{abstract}
\vspace*{-0.4cm}

Image-based lighting (IBL) is a widely used technique that renders objects using a high dynamic range image or environment map.
However, aggregating the irradiance at the object's surface is computationally expensive, in particular for non-opaque, translucent materials that require volumetric rendering techniques.
In this paper we present a fast neural 3D reconstruction and relighting model that extends volumetric implicit models such as neural radiance fields to be relightable using IBL. It is general enough to handle materials that exhibit complex light transport effects, such as translucency and glossy reflections from detailed surface geometry,
producing realistic and compelling results.
Rendering can be within a second at 800$\times$800 resolution (0.72s on an NVIDIA 3090 GPU and 0.30s on an A100 GPU) without engineering optimization.
Our code and dataset are available at \url{https://zhusz.github.io/TRHM-Webpage/}.
\end{abstract}

\begin{figure*}[!b]
\begin{center}
\centerline{\includegraphics[width=\textwidth]{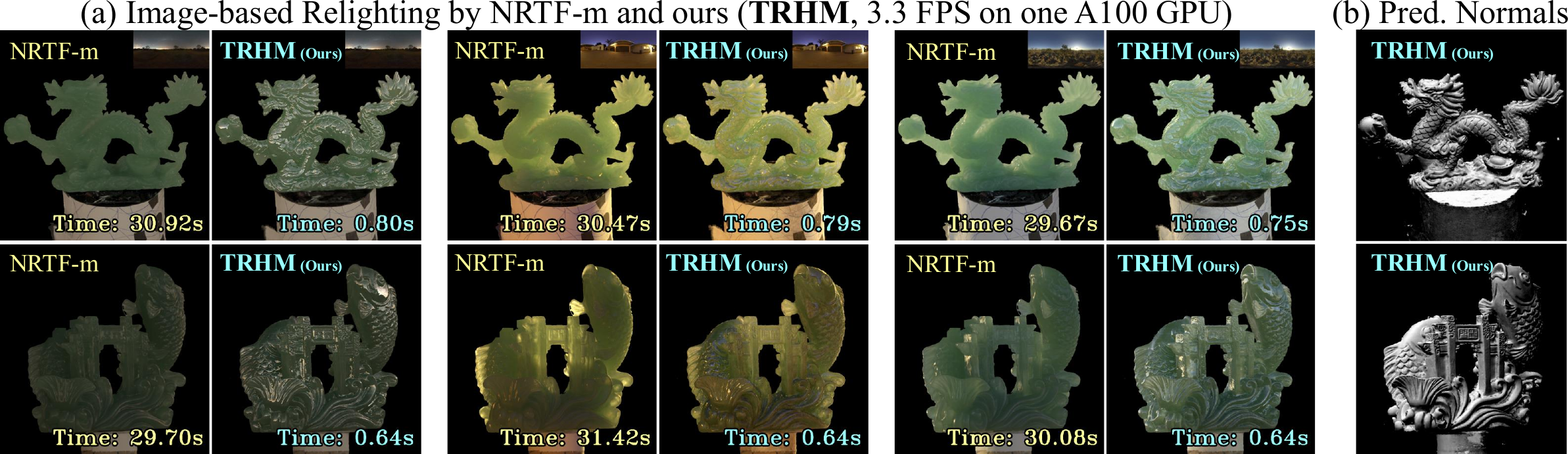}}
\vspace*{-0.3cm}
\caption{Rendering pipeline of the proposed \textbf{TRHM} model. The testing (rendering) undergoes a low-frequency branch (Sec.~\ref{sec:method:lowfreq}) with a hyper-net predicting based on a low-resolution envmap; and a high-frequency branch (Sec.~\ref{sec:method:reflections}) where dedicated reflection normal $\hat{n}''$ (Sec.~\ref{sec:method:normals}) is estimated for facilitating all the reflective high-frequency effects. Only a single rendering pass is necessary for full, image-based illumination. Our training (optimization) undergoes three stages that can effectively transfer from the OLAT data to the envmap prediction mode. }
\label{fig:teaser}
\end{center}
\vspace{-0.4in}
\end{figure*}

\section{Introduction}
\label{sec:intro}

Image-based lighting encodes the illumination at a point in space from all directions in an image, the environment map, and uses this information to realistically light an object.
It is a widely used technique in industry due to its flexibility: natural illumination from the physical world can swiftly be recorded using a 360 degree camera to create a light probe and then used to place virtual objects in real scenes seamlessly.
Depending on its resolution, such a light probe is sufficient to describe the entire far field environmental light, and can be used to render all kinds of materials.
Rough and glossy surfaces can be reproduced correctly and may contain a high resolution reflection of the environment, for example; for translucent objects, the entire light transport can be computed from all directions.
However, that comes at a cost: if the light probe is used directly, even for a low resolution of 16$\times$32, 512 passes would have to be performed if no simplifying assumptions are being made.
Even for highly optimized rendering frameworks this makes rendering a single frame too slow for real-time applications, unless simplifying assumptions about the materials and lighting are made (e.g. pre-convolving the environment for high-roughness materials, often used in real time rendering).
Yet, combining IBL with radiance field reconstruction techniques and pre-computed radiance transfer (PRT) is highly attractive.
The combination of these techniques covers the radiance transfer through an object in its entirety without any simplifying assumptions and allows rendering of even the most complex materials faithfully.

In this paper, we propose a method to achieve this at sub-second performance for rendering an 800 $\times$ 800 frame. The runtime grows linearly with the resolution, without making particular assumptions on the envmap resolution.
It accounts for various lighting effects, ranging from subsurface scattering to specular highlights on glossy surfaces.  
Our model avoids repeatedly querying the environment for accumulation and instead uses a neural model to `summarize' them.
This works well for low frequency illumination and has been explored in the past with HyperNets.
However, this approach does not fare well for high frequency details: for this scenario, we propose a separate model stream that uses a reflection hint pyramid that can be precomputed and is queried only once given the computed normal direction.
This means, our model achieves IBL for low- and high-frequency lighting including reflections in a single pass.
We name the full model TRHM (Translucent-Reflection Hybrid Modeling).

\begin{table*}[t]
\hspace{-0.15in}
\resizebox{\textwidth}{!}{
\normalsize
\begin{tabular}{lccccc}
\Xhline{4\arrayrulewidth}
                                                 &    \citet{zhang2022iron}  & \citet{deng2022reconstructing} & \citet{lyu2022neural} & TRHM \\ \hline
Fast Image-Based Relighting                               &               {\color{blue} \cmark}                                       &                  {\color{red} \xmark}                                      &                         {\color{red} \xmark}                                            &   {\color{blue} \cmark}   \\
Topology Flexibility                                 &           {\color{blue} \cmark}                                            &                {\color{red} \xmark}                            &        {\color{blue} \cmark}                                          & {\color{blue} \cmark}                     \\
Subsurface Scattering                                &                           {\color{red} \xmark}                         &                     {\color{blue} \cmark}                                     &       {\color{blue} \cmark}                                                       &           {\color{blue} \cmark}    \\ 
Local Micro-Geometry                        &                              {\color{red} \xmark}                       &                     {\color{red} \xmark}                             &                                             {\color{red} \xmark}             &                        {\color{blue} \cmark}                                     \\ 
\Xhline{4\arrayrulewidth}
\end{tabular}}
\vspace{-0.1in}
\caption{A comparison with recent state-of-the-art relighting approaches. Our approach is fully data-driven, making no particular assumptions on material properties (e.g. opaqueness, lambertian, ...). We optimize the implicit density geometry with the appearance end-to-end, demonstrating topology flexibility and independence of geometry initialization during optimization. We maintain a wide range of lighting effects, such as subsurface scattering and local micro-geometry modeling. Our algorithm requires runtime of less than a second when given an environment map as the lighting input. It is worth pointing out that the full NRTF model~\citep{lyu2022neural} cannot handle subsurface scattering. In the table we report NRTF-m~\citep{lyu2022neural} instead. }
\vspace{-0.15in}
\end{table*}

Reconstructing an object at this level of fidelity requires high quality ground truth data, captured using a light stage.
Our approach avoids relying on a large collection of image captures under large quantities of environment maps, but instead, is able to learn from an OLAT collection - a common and reasonable efforts of data collections in practice.
We have recorded several real-world objects featuring subsurface scattering in a light stage, and show that our method produces high quality visual results in recorded as well as in novel lighting conditions.
These objects have been captured with high fidelity, featuring rich, high frequency spatially-varying details, resulting in 15~TiB of data, which is 3000 times larger and notably more detailed than current data for research in this area~\citep{deng2022reconstructing}.
We will release both, our training framework and the dataset upon publication.

Our main contributions are as follows:
\begin{itemize}
    \item We present a novel reconstruction and rendering framework that uses image-based lighting together with pre-computed radiance transfer and volumetric neural rendering techniques to create high fidelity reconstructions and renderings of objects recorded in a light stage.
    \item To the best of our knowledge, this approach is the first one to achieve sub-second rendering speeds in this setting.
    \item Our model captures local micro-geometries from the image collections and provides faithful lighting effects associated with these captured geometric details. Our results indicate our approach demonstrate advantages compared to existing methods with respect to pixels related to specular highlights and micro geometries.
    \item Our approach enables image-based relighting only from light-stage captures (\textbf{OLAT} data), bypassing the necessity to capture a large collection of images under a large number of \textbf{envmaps} that is generally hard to obtain.
    
\end{itemize}

Our code and dataset are available at \url{https://zhusz.github.io/TRHM-Webpage/}.

\section{Related Work}
\label{sec:related}

\textbf{Relighting and Surface Representations.} 
The problem of relighting an object or a scene under novel lighting conditions has been extensively studied. 
Usually, the problem is tackled via decomposing the appearance into the lighting and the surface material properties. 
Early works estimate material given known illumination such as a single light source~\citep{yu1999inverse,debevec2000acquiring} or spherical gradient illumination~\citep{fyffe2009cosine,guo2019relightables} with known geometry. 
\citet{zhang2021nerfactor} directly model light transports with known illuminations and known geometry. 
More recently, neural scene representations~\citep{xie2022neural} and differentiable rendering~\citep{nimier2019mitsuba} allow us to jointly optimize BRDF and geometry. 
Some methods apply inverse rendering using implicit surface to obtain materials~\citep{luan2021unified,zhang2022iron,munkberg2022extracting}. Other approaches utilize volumetric representations with opacity fields~\citep{bi2020deep,bi2020neural,zhang2021physg,zhang2021nerfactor,boss2021nerd}. The required illumination setup can be reduced to a co-located light~\citep{bi2020deep,bi2020neural}, and unknown illuminations~\citep{luan2021unified,zhang2022iron,zhang2021physg,zhang2021nerfactor,boss2021nerd}. To reduce the ambiguity in BRDF, the aforementioned methods use parametric BRDFs such as a microfacet model~\citep{walter2007microfacet,burley2012physically}. However, these parametric models do not support subsurface scattering as they only consider reflectance. In contrast, our approach deals with global light transport effects including subsurface scattering.

\textbf{Subsurface Scattering.} Subsurface scattering refers to light transport inside of a translucent object. 
It is quite common in the real world, being characteristic of many types of materials, such as wax, jade, soap, fruits, skin, to name a few.
Radiance at any point of the object's surface is not only the result of direct reflection from the light hitting such point, but also many light paths coming from different parts of the object. 
Thus, surface representations (e.g. various BRDFs) cannot represent this type of light transmission.
While subsurface scattering can be accurately modeled by volumetric path tracing algorithms~\citep{novak2018monte}, their run time is typically prohibitive in certain applications, despite efforts to accelerate brute-force computation, e.g.~through a shape adaptive learned SSS model~\citep{vicini2019learned} that relies on a conditional variational auto encoder that learns to sample from a distribution of exit points on the object surface.
Some other works have focused on estimating the scattering parameters from images of translucent objects. 
Inverse Transport Networks~\citep{che2020towards} infer the optical properties that control subsurface scattering inside translucent objects of any shape under any illumination, using a physically-based differentiable path tracer.
Another approach based on stochastic gradient descent, combined with Monte Carlo rendering and a material dictionary, was capable of estimating the scattering materials, inverting the radiative transfer parameters~\citep{gkioulekas2013inverse}.
Nevertheless, since volumetric path tracing can be costly, applying a BSSRDF can be a faster alternative~\citep{deng2022reconstructing}. 
Compared to BRDF-based representations, a higher dimension of inputs (usually 6D for homogeneous materials) is fed to query the outgoing radiance. 
A relighting algorithm can thus seek to optimize the BSSRDF function with the inverse rendering process so that the resulting material can be relit in conventional rendering engines.
Our work follows a different path - we learn our relighting model in a fully data driven fashion, and learn the cached outgoing radiance for each point using a deep neural network, where we bypass the expensive BSSRDF computation in our optimization iteration.
We will release high-resolution and large scale light stage dataset with rich lighting effects, such as translucency coupled with specular highlights and translucent shadowing, facilitating future research.

\textbf{Neural Radiance Fields and Precomputed Radiance Transfer.} 
Neural Radiance Fields (NeRF)~\citep{mildenhall2020nerf, barron2021mip, barron2022mip} optimize a parameterized volume rendering model from multiple views of the scene so that at test time, novel views can be synthesized from the learned model.
Despite its superior rendering quality, NeRF bakes all the lighting and reflective surface information into the RGBs without modeling the interaction of the light and the material.
Recent studies~\citep{lyu2022neural} have shown promising results for relightable models via incorporating the idea of ``precomputed radiance transfer'' (PRT)~\citep{sloan2002precomputed} from the real time rendering community.
Instead of precomputing and caching the intermediate representation per location, they seek to optimize a cached intermediate representation in the reconstruction process. 
Notably, \citet{lyu2022neural} relies on a fairly accurate pre-computed surface~\citep{wang2021neus}, and keeps the lighting appearance optimization separate from the geometry acquisition process.
Focused on synthetic images with varying but known illumination, a NeRF extension~\citep{zheng2021neural} was presented to reconstruct participating media with full global illumination effects. 
\citet{yu2023learning} addresses relighting with translucent objects using scattering functions.
In contrast, our novel volume rendering framework not only enables optimizing the geometry details with appearance cues, but also works on scenes with partially opaque mass (e.g. thin rope or furs) and demonstrates high quality results on synthetic and real data.
In addition to distant point lights, our approach efficiently relights the captured scenes with environment maps by using the hyper-network~\citep{ha2017hypernetworks,bi2021deep,iwase2023relightablehands}, avoiding prohibitively expensive evaluation of each environment map pixel independently.

\vspace{-0.2in}
\section{Neural Image-based Relighting}
\label{sec:method}

\begin{figure*}
\begin{center}
\centerline{\includegraphics[width=\textwidth]{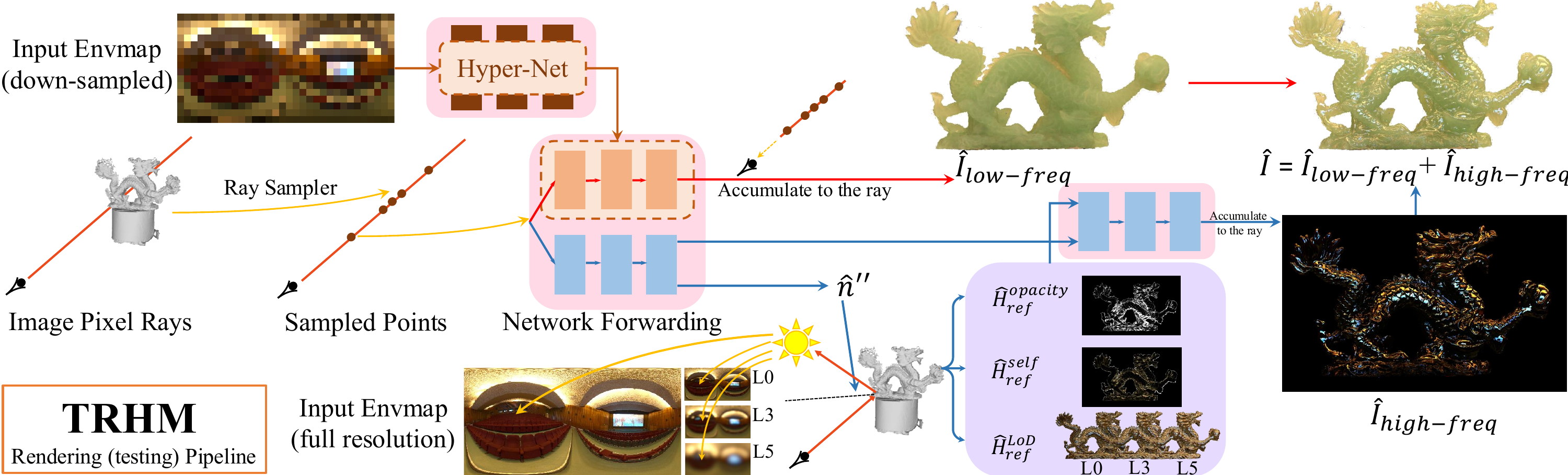}}
\vspace*{-0.3cm}
\caption{Rendering pipeline of the proposed \textbf{TRHM} model. The testing (rendering) undergoes a low-frequency branch (Sec.~\ref{sec:method:lowfreq}) with a hyper-net predicting based on a low-resolution envmap; and a high-frequency branch (Sec.~\ref{sec:method:reflections}) where dedicated reflection normal $\hat{n}''$ (Sec.~\ref{sec:method:normals}) is estimated for facilitating all the reflective high-frequency effects. Only a single rendering pass is necessary for full, image-based illumination. Our training (optimization) undergoes three stages that can effectively transfer from the OLAT data to the envmap prediction mode. }
\label{fig:overview}
\end{center}
\vspace{-0.4in}
\end{figure*}

\vspace{-0.1in}
\subsection{Overview} 
\label{sec:method:overview}
\vspace{-0.1in}

Our approach relights a scene based on the given query lighting input - an environment map that is denoted as $\mathbf{E} \in \mathbb{R}^{M_E \times N_E \times 3}$, where $M_E$ and $N_E$ denotes the height and width of $\mathbf{E}$.
Our output is the HDR image rendering of the scene $\{\hat{\mathbf{I}}|\hat{\mathbf{I}} \in \mathbb{R}^{M_I \times N_I \times 3}\}$ under the given lighting envmap $\mathbf{E}$.
We denote each pixel ray as $\mathbf{r} \in \mathbb{R}^3$, with its radiance from the image capture as $\hat{I}(\mathbf{r}) = \hat{I}(x, y)$, where $(x, y)$ are the pixel coordinates of the ray $\mathbf{r}$ on the image plane. 
We thus denote our prediction problem as $\hat{I}(\mathbf{r}; \mathbf{E})$. 
While we primarily aim to predict the relighting results based on the envmap-based lighting condition, our training data consists of the point-light-based light stage image captures, as the most efficient and viable format to capture all possible lighting effects of the object scenes.
During training, our model is optimized from a collection of HDR images $\{\mathbf{I} | \mathbf{I} \in \mathbb{R}^{H_{I} \times W_{I} \times 3}\}$ that capture the scene from multiple views, with each image associated with the camera poses as well as the point light information. 
We denote the incident direction of the point light as $\omega_l \in \mathbb{R}^3$, and the prediction can thus be conditioned on the point light as $\hat{I}(\mathbf{r}; \omega_l)$.
Throughout the methodology section, all the predicted variables are denoted with ``~$\hat{~}$~''.

\textbf{Assumptions.} We assume the given image-based lighting is not a near-field lighting as a result of using environment maps as inputs.
We assume all the subsurface scattering effects are isotropic.  
The hard shadow effect is not explicitly modeled in our approach and is beyond the scope of this work. 
We handle indirect lighting up to one bounce because we saw most of the energy comes from light paths no more than one bounce for practical purposes. 

\textbf{Modeling.} We propose our fast neural relighting algorithm for image based lighting that faithfully captures lighting effects, such as translucency, subsurface scattering as well as hard shadows, specular highlights and glossy reflections. 
We term our model as translucent-reflection hybrid modeling (TRHM), that stands for the two most representative lighting effects. 
With an environment map as the input, we do not do expensive enumeration of every pixel as a point light to aggregate the rendering results~\citep{lyu2022neural}.
Instead, we propose a set of techniques for enabling image-based lighting without querying the same lighting network numerous times, including \textit{i)} a hypernet-based~\citep{ha2017hypernetworks,bi2021deep,iwase2023relightablehands} distillation of the neural radiance transfer field for low-frequency effects such as diffuse component and subsurface scattering; \textit{ii)} learning of the reflection effects among various surface roughnesses with an irradiance level of details (LoD) of the lighting envmap; and \textit{iii)} learning of the local micro-geometry with the help of the specular highlight cues. 
By combining both the low-frequency (the hyper-net) and high-frequency (reflections) prediction, our model demonstrates a wide range of lighting effects under various lighting inputs. 

\textbf{Optimization (training) pipeline overview.} Our model demonstrates the capability of transferring from the optimization of the point-light based data (e.g. a light stage) into the image-based lighting prediction at rendering time. 
The key is the compatibility of our proposed LoD-based reflection prediction hints for both the point-light and the environment map prediction mode. 
Our learning framework undergoes three steps for the transferring.
In the first step (Sec.~\ref{sec:method:normals}), we devise the point-light-based relighting optimization scheme to separate low and high frequency lighting effects and capture the local micro geometry of the surface, guided by the pixel loss (Eq.~\ref{eq:losspixel}), the surface normal losses (Eq.~\ref{eq:lossdotproduct}-\ref{eq:lossnewtie}) and the high-frequency regularization loss for separating the low and high frequency predictions (Eq.~\ref{eq:losshighfreqreg}).
In the second step (Sec.~\ref{sec:method:reflections}), we optimize with the envmap lighting condition guided by the levels-of-details (LoD) of the envmap, with the normal fixed while still enabling the pixel loss (Eq.~\ref{eq:losspixel}) and the high-frequency regularization loss (Eq.~\ref{eq:losshighfreqreg}).
In the last step (Sec.~\ref{sec:method:lowfreq}), we optimize a fast hyper-net envmap-based low-frequency lighting effects prediction model via distillation.
Figure~\ref{fig:overview} illustrates our rendering (testing) pipeline. 

\begin{figure*}
\begin{center}
\centerline{\includegraphics[width=1\textwidth]{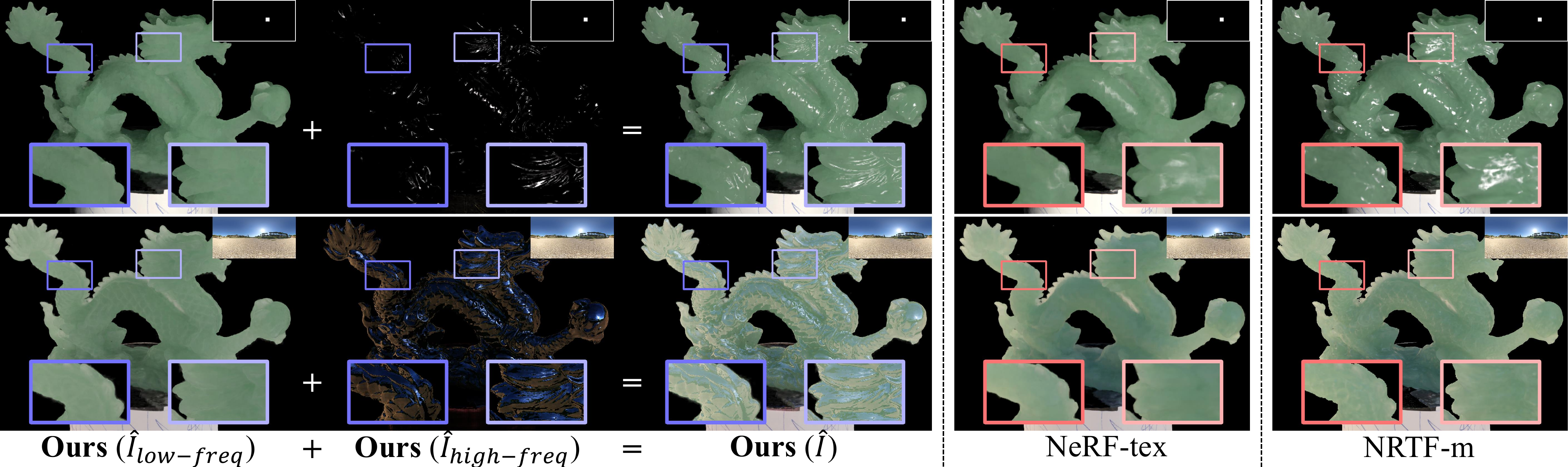}}
\vspace*{-0.3cm}
\caption{Separation into low-frequency and high-frequency (top: for point light; bottom: for an envmap) enables the proposed framework to render complex materials with high fidelity. In this example, the jade structure is present in the low frequency rendering, but does not exhibit specular highlights. These are captured well in the high frequency component, leading to a faithful rendering. Without the separation, reconstruction of this material is not possible. Best viewed electronically with zoom-in.}
\label{fig:lowhigh}
\end{center}
\vspace{-0.4in}
\end{figure*}

\vspace{-0.1in}
\subsection{Step 1 - Point-light-based Relighting with Local Micro Geometry Modeling}
\label{sec:method:normals}
\vspace{-0.1in}

\textbf{Point-light-based Base Model}. While we primarily address the problem of relighting under the image-based lighting condition (an input envmap), our first step starts from training with the OLAT data where the point light is the given lighting condition.
By incorporating the radiance transfer modeling into the neural relighting framework~\citep{lyu2022neural}, one can enable global illumination and subsurface scattering with orthogonal basis lighting representations. 
For point lights, adding the additional dimensions that encode the incident lighting direction $\omega_l$~\citep{lyu2022neural,yu2023learning,baatz2022nerf} would achieve global illumination effects,
i.e. for each query point $\mathbf{x} \in \mathbb{R}^3$ and the associated viewing direction $\mathbf{d} \in \mathbb{R}^3$, the color prediction becomes $\hat{c}(\mathbf{x}; \mathbf{d}; \omega_l)$, with the additional input $\omega_l$ of the point light information.
The predicted pixel color $\hat{I}(\mathbf{r})$ could be either the volume rendering accumulation of the predicted per-point color $\hat{c}$~\citep{baatz2022nerf, yu2023learning} or the single predicted color $\hat{c}$ at the depth computed by z-buffer~\citep{lyu2022neural}.
In our practice, we apply the volume rendering scheme and the density of the neural rendering model~\citep{mildenhall2020nerf} is finetuned from a pre-optimized Ref-NeRF model~\citep{verbin2022ref}.

\textbf{Separating the Low and High Frequency Predictions}. It is worth pointing out that while such additionally incorporated point light information $\omega_l$ could provide plausible relighting results even for high frequency lighting effects, 
our experimentation favors a separation of the low and high frequency lighting effects modeling that provides the good opportunity to transfer the point-light-based prediction to the envmap-based prediction, i.e.
\begin{equation}
\vspace{-0.05in}
\hat{I}({\textbf{r}}) = \hat{I}_\text{low-freq}(\textbf{r}) + \hat{I}_\text{high-freq}(\textbf{r}).
    \label{eq:full}
\end{equation}
Separating the prediction into low and high frequencies brings several key benefits. 
First, converting the point-light-based model into the the hyper-net-based model in Step 3 (Sec.~\ref{sec:method:lowfreq}) is insufficient to capture all the high-frequency lighting effects given an envmap, due to its limitation of the representation dimension of the input envmap.
Second, our conversion from the point-light-based to the envmap-based learning could apply different algorithms to different branches of the pixel prediction (Sec.~\ref{sec:method:reflections}). 
Lastly, as discussed next, we could put particular local-micro geometry modeling solely with the high frequency branch prediction. 
Figure~\ref{fig:lowhigh} provides the illustration of our prediction from both frequency branches.

\textbf{Our Point-light-based Modeling with local micro-geometry modeling}. We noticed that the quality of modeling of the specular highlights as well as glossy reflections is highly sensitive to the accurate modeling of the complicated local micro geometry over the object surface, particularly for materials with very small roughness.  
A tiny fluctuation on the surface would significantly alter the specular highlight intensities of the related pixels.
Learning without taking the local geometry into account leads to an over-smoothed pixel intensity prediction as the ground truth HDR values tend to naively average with each other with the highlight on or off (e.g. the point light version of NeRF-tex~\citep{baatz2022nerf} in Fig.~\ref{fig:lowhigh}). 
We found that accurately predicting the normals over the object surface can significantly enhance the accuracy of the highlight and reflection prediction. 
Motivated by \citet{verbin2022ref}, where a separately estimated surface normal is used for computing the reflection direction encoding for avoiding linking with the highly noisy density gradient, we propose to learn an additional normal prediction that is solely for the reflection purpose.

\begin{figure*}
\begin{center}
\centerline{\includegraphics[width=0.7\textwidth]{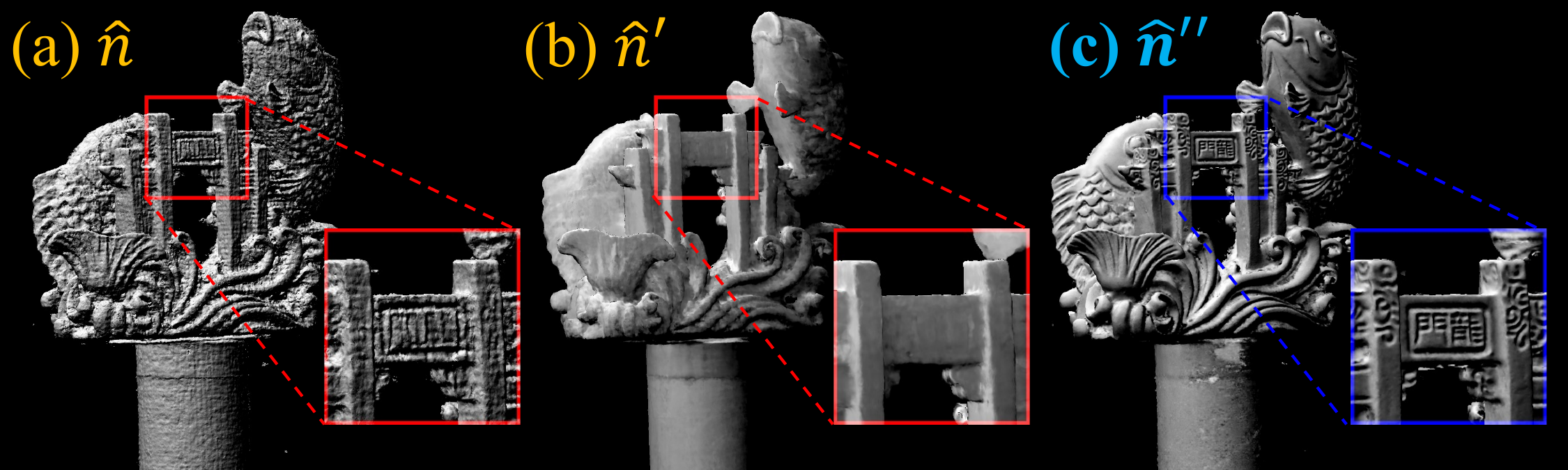}}
\vspace*{-0.2cm}
\caption{Our model prevails on capturing local micro-geometry details with appealing accuracy of surface normal prediction (c, $\hat{n}''$). Here we visualize our predicted normal for the reflection purpose and compare with the noisy analytical gradient normal (a, $\hat{n}$) as well as the overly-smoothed RefNeRF-predicted normal (b, $\hat{n}'$) (originally proposed for Integrated Direction Encoding~\citep{verbin2022ref} rather than reflections).}
\label{fig:normal}
\end{center}
\vspace{-0.4in}
\end{figure*}

Given the set of captured images under the point light training data that could precisely separate the reflection effects from only a single light source, we seek to optimize the separately predicted normal, denoted as $\hat{\textbf{n}}''(\textbf{r})$, on top of the two existing normals in the Ref-NeRF framework~\citep{verbin2022ref}, namely the ``density gradient normal'' ($\hat{\textbf{n}}(\textbf{r})$) and the ``Ref-NeRF predicted normal'' ($\hat{\textbf{n}}'(\textbf{r})$).
We name this new normal $\hat{\textbf{n}}''(\textbf{r})$ as ``predicted normal for reflection purpose only''.
The normal $\hat{\textbf{n}}''$ is associated with the ray $\textbf{r}$ and is obtained via volume rendering accumulation of the normal predictions along $\textbf{r}$. 
As we found, under the Ref-NeRF density modeling framework, the density is highly concentrated near the surface~\citep{verbin2022ref}, and the obtained normal $\hat{\textbf{n}}''(\textbf{r})$ is close enough for representing the predicted normal on the surface.
We use a simple Blinn-Phong shading model to predict the specular highlight via 
\begin{equation}
\vspace{-0.05in}
    \hat{I}_\text{high-freq}(\textbf{r}) = \hat{A}(\textbf{r}) ~ \text{Dot}(\hat{\textbf{n}}''(\textbf{r}), \textbf{h}(\textbf{r}))^{\hat{\alpha}(\textbf{r})},
\vspace{-0.05in}
\end{equation}
where both $\hat{A}(\textbf{r})$ and $\hat{\alpha}(\textbf{r})$ are predicted per-point and accumulated along the ray $\textbf{r}$ similar to $\hat{\textbf{n}}''(\textbf{r})$.
$\textbf{h}(\textbf{r})$ is the half direction between the lighting direction and the viewing direction as used in the Blinn-Phong model.
To train this new normal $\hat{\textbf{n}}''(\textbf{r})$, we impose three loss terms to acquire the high quality prediction.
First, we compute the final prediction $\hat{I}(\textbf{r})$ in the same way as in Eq.~\ref{eq:full}, and then impose the pixel loss
\begin{equation}
\vspace{-0.05in}
    L_\text{pixel}(\textbf{r}) = \| \hat{I}(\textbf{r}) - I(\textbf{r}) \|_2^2
    \label{eq:losspixel}
\end{equation}
Second, by collecting a small set of pixels that demonstrate extremely high radiance intensities that is saturated by our cameras, we directly impose the dot-product loss on these rays:
\begin{equation}
\vspace{-0.05in}
    \begin{aligned}
        L_\text{dot-product} = \text{ReLU}(\|I(\textbf{r})\|_1 - I_0) \cdot ( 1 - \text{Dot}(\hat{\textbf{n}}''(\textbf{r}), \textbf{h}(\textbf{r})) ),
    \end{aligned}
    \label{eq:lossdotproduct}
\end{equation}
where $\text{ReLU}(\|I(\textbf{r})\|_1 - I_0)$ is a weighting factor and $I_0$ is a constant threshold for clipping out non-highlight pixels.
Lastly, we regularize our predicted normal $\hat{\textbf{n}}''(\textbf{r})$ by letting it tie to the Ref-NeRF predicted normal $\hat{\textbf{n}}'(\textbf{r})$, 
\begin{equation}
\vspace{-0.05in}
    L_\text{tie-ref} = \|\hat{\textbf{n}}''(\textbf{r}) - \hat{\textbf{n}}'(\textbf{r}) \|^2_2,
    \label{eq:lossnewtie}
\vspace{-0.05in}
\end{equation}
to ensure $\hat{\textbf{n}}''(\textbf{r})$ is not deviating too far.

Our predicted reflection normals, $\hat{\textbf{n}}''(\textbf{r})$, demonstrate superior quality compared to alternative candidates  (e.g. $\hat{\textbf{n}}(\textbf{r})$ and $\hat{\textbf{n}}'(\textbf{r})$). 

Specifically, $\hat{\textbf{n}}''(\textbf{r})$ avoids the heavy dependency on density gradients inherent to $\hat{\textbf{n}}(\textbf{r})$, while preserving micro-geometric details that are often lost in the overly smoothed predictions of $\hat{\textbf{n}}'(\textbf{r})$~\citep{verbin2022ref} (Fig.~\ref{fig:normal}). 
This enhanced geometric fidelity enables our model to predict high-quality specular reflections over glossy surfaces. 

For better separating the low-frequency and high-frequency intensities during training, we explicitly cap the maximum value of $\hat{I}_\text{low-freq}$ using a sigmoid activation function, and also incorporate an additional loss that penalizes the HDR intensity value of the high-frequency prediction, i.e. 
\begin{equation}
\vspace{-0.05in}
    L_\text{high-freq-reg} = \|\hat{I}_\text{high-freq}\|_2^2.
\label{eq:losshighfreqreg}
\vspace{-0.05in}
\end{equation}

By optimizing with the weighted sum of the loss function in Eq.~\ref{eq:losspixel}-\ref{eq:losshighfreqreg}, our point-light-based relighting model is better ready for transferring to the envmap-based lighting condition compared to the base point-light-based model.

\subsection{Step 2 - Transferring to the Image-based Relighting with Reflection Hints}
\label{sec:method:reflections}

Under the point light condition, Zeng et al.~\citep{zeng2023relighting} proposed to incorporate the reflection hints, a GGX micro-facet representation with four pre-defined roughnesses, for predicting specular highlights. 
When using an environment map, the lighting becomes a large collection of light sources, and it is difficult to efficiently fit to the above-mentioned representation without the enumerate-and-aggregate inference process.
We observe that the reflection of a point on the surface is highly dependent on the envmap intensities in the reflection direction as well as its adjacent directions, supposing no occluder is present on the reflection ray.
With a lower roughness value, the reflection is impacted by a narrower range of envmap directions, while a higher roughness value means a wider range of envmap directions impacts reflection value.
We propose to utilize a pyramid of envmaps, denoted as $\{\mathbf{E}_j\}_{j=0, 1, 2, ...}$, with each level $l$ processed by a Gaussian filter of a particular kernel size and standard deviation, to represent the levels of details (LoD) of the envmap and approximate the reflection hints for various surface roughnesses. 
We simply query the envmap HDR values $\textbf{E}_j(\hat{\omega}_r)$ in the predicted reflection direction $\hat{\omega}_r$~\citep{verbin2022ref} at each level $j$, and concatenate them to obtain our `reflection hints' under the image-based lighting condition:
\begin{equation}
\vspace{-0.05in}
    \hat{H}_\text{ref}^\text{LoD} = \{\mathbf{E}_j(\hat{\omega}_r)\}_{j=0, 1, 2, ...}.
\vspace{-0.05in}
\end{equation}

\begin{figure*}
\begin{center}
\centerline{\includegraphics[width=0.7\textwidth]{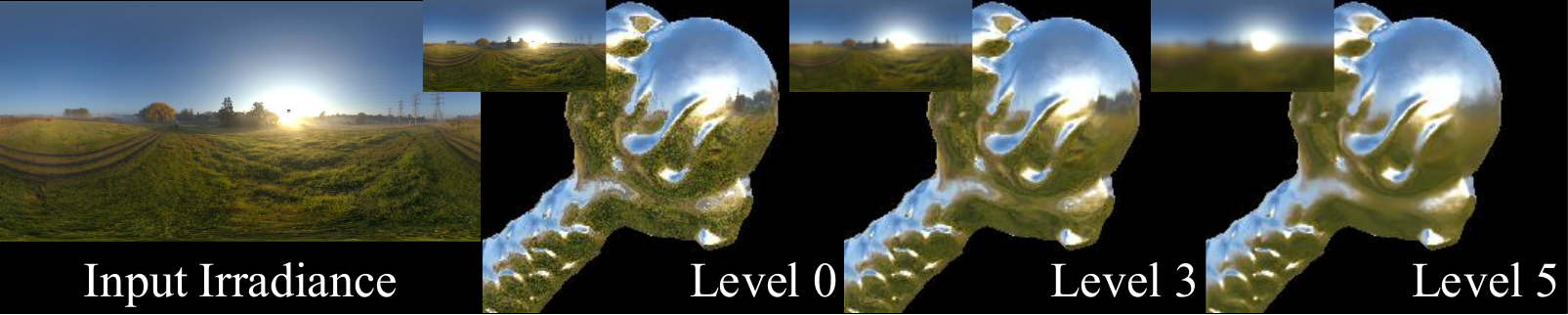}}
\vspace*{-0.2cm}
\caption{Reflection hint pyramid visualization ($\hat{H}_\text{ref}^\text{LoD}$). The reflection hints model reflected irradiance and take self occlusion into account. As the pyramid level increases, roughness increases and the reflection becomes more blurry. Note they are the input to our high frequency branch rather than our outputs of roughness progression (Fig.~1). The mini irradiance map on the top-left corner of each visualized level (0, 3, 5) shows the filtered input envmap by the gaussian filter.}
\label{fig:refhints}
\end{center}
\vspace{-0.4in}
\end{figure*}

Fig.~\ref{fig:refhints} illustrates the reflection hints based on the given environment map lighting, where we simply cast the object material as the mirror for visualization purpose.
The hints $\hat{H}_\text{ref}^\text{LoD}$ retrieved from different levels of the environment map demonstrate different levels of blurriness of the environment map, serving for hints for different levels of roughness of the materials of the surface reflection. 
The hint serves a similar functionality as the point-light reflection hint~\citep{zeng2023relighting}, with the difference that the point-light reflection hint~\citep{zeng2023relighting} directs the model to interpolate between the GGX reflection response under varying surface roughness, while our proposed image-based reflection hint instead lets the model to interpolate between the varying levels of the envmap HDR response. %
On one hand, our hint relaxes the constraint of using only a single point-light, and can be seamlessly integrated into the image-based lighting prediction framework. 
On the other hand, the choice of this hint further enables the flexibility of transferring the learning from the point-light training data (e.g. a light stage) to the image-based lighting prediction scenario, where different levels of material roughness shall be able to receive point light signal if the reflection direction $\hat{\omega}_r$ is close to, but no identical to, the point light direction $\omega_l$.

\textbf{Handling of self-occlusions.}
To accommodate self-occlusions along the reflection direction as well as approximating the prediction of the reflected light for more than one surface bounce, we further incorporate the self-occlusion cue as part of our reflection hint.
More precisely, we further incorporate the opacity prediction as well as the incident radiance as part of our reflection hint.
We predict these hints via putting a virtual eye on the predicted surface point, looking toward the predicted reflection direction of the ray ($\hat{\omega}_r(\mathbf{r})$).
We denote the obtained opacities as well as the radiance color along the reflection direction as $\hat{H}_\text{ref}^\text{opacities} \in \mathbb{R}$ and $\hat{H}_\text{ref}^\text{self} \in \mathbb{R}^3$ respectively.
Note that when predicting $\hat{H}_\text{ref}^\text{self}$, it further requires to trace to the next reflection bounce, resulting in a recursive problem. We found that setting $\hat{H}_\text{ref}^\text{incident}$ to zero when computing the hints bypasses this issue while maintaining promising visual results.
The full version of our proposed reflection hint could be written as 
\begin{equation}
\vspace{-0.05in}
    \hat{H}_\text{ref} = \{ \hat{H}_\text{ref}^\text{LoD}, \hat{H}_\text{ref}^\text{opacities}, \hat{H}_\text{ref}^\text{self} \} .
\vspace{-0.05in}
\end{equation}
Based on this hint, we devised a 3-layer MLP to predict the high-frequency prediction $\hat{L}_\text{high-freq}(\mathbf{r})$, and compute the full prediction via Eq.~\ref{eq:full}.
For optimization, we fix all the weights for predicting density and reflection normal $\hat{\mathbf{n}}''$, and tuning the remaining model weights with only the loss function Eq.~\ref{eq:losspixel} and Eq.~\ref{eq:losshighfreqreg}. 
All the pixel intensity predictions are purely from the output of the MLPs, and the Blinn-Phong shading model used in the previous step (Sec.~\ref{sec:method:normals}) are discarded in this step - only the obtained reflection normal $\hat{\mathbf{n}}''$ is retained from the previous step.
After the optimization procedure of this step, we further distill the current Ref-NeRF-based~\citep{verbin2022ref} model into the Hash-Volume Grid~\citep{muller2022instant} w.r.t. the density modeling to achieve sub-second testing speed.

\subsection{Step 3 - Hypernet-based Radiance Transfer Fields}  
\label{sec:method:lowfreq}

Our relighting model after Step 2 could predict with the image-based envmap input only for its high-frequency branch. 
To fully enable envmap-based prediction (particularly, the low-frequency branch), we propose to use a hyper network~\citep{ha2017hypernetworks} to generate the parameters of the color branch of the NeRF network ($\hat{\Theta}$) for better representational power, motivated by recent related approaches~\citep{bi2021deep,iwase2023relightablehands}.
Our hypernet-based prediction thus becomes
\begin{equation}
\vspace{-0.15in}
    \begin{aligned}
        \hat{\Theta}_\text{color} & = \mathcal{H}(\mathbf{E}), ~~~
        \hat{I}_\text{low-freq} & = f_{\hat{\Theta}}(\mathbf{x}; \mathbf{d}).
    \end{aligned}
\vspace{-0.05in}
\end{equation}
We use a downsampled low resolution version of the input HDR envmap $\mathbf{E}$ (16 $\times$ 32) as input to the hyper network for the ease of learning the hyper-net with the finite set of envmaps during training.
We empirically found that a 3-layer MLP is sufficient for including the lighting information to the NeRF network.
During optimization, the only loss term is Eq.~\ref{eq:losspixel} and we fix the whole high-frequency branch in this step.

\section{Experiments}

\subsection{Data}

\noindent \textbf{Image-Based Lighting Data}. To provide the distribution of the evnrionment map as the lighting condition as used in our experiments, we utilize the lighting data from the Laval dataset~\citep{gardner2017learning}, in which a total of 2233 high quality indoor HDR environment maps are provided. 
We utilize a fixed set of 2000 HDR maps used in training, and the remaining 233 for testing or visualization.
During training, to provide a reliable and sufficiently diverse HDR condition, we augment the 2000 HDR environment map in the same way as in \citep{bi2020deep}. 
Our augmentation is conducted on the fly with the training iterations and hence each augmentation is unique. All of them shall be used for distilling the OLAT-based low-frequency relighting model into the envmap-based (Sec.~\ref{sec:method:lowfreq}).

\noindent \textbf{Light Stage Capture Scenes: the TOLC dataset} (the Translucent-OLAT-Lightstage-Capture dataset). 
We captured and released a large and high-resolution light stage dataset, consisting of 11 real object scenes on their appearance under OLAT setting, for evaluating the effects of various approaches.
Our light stage data contains 9 groups and 20 cameras per scene (a total of 180 views), with each view consisting of 331 OLAT renderings, thus leading to a total of 59580 HDR images per scene. 
Among all the captured scenes, two of them (jade-dragon and jade-fish) demonstrates strong specular effects while we also include another six scenes without major high-frequency effects. 
During training, for each scene, we use the first 18 cameras in each group, and use 75 out of the 331 OLATs for training, leading to a total of 12150 training images per scene. 
During testing we only include samples with both, unseen lighting directions and unseen views. 
For the OLAT-based evaluation, we use the remaining 2 cameras from each group (a total of 18 views) and 10 unseen OLATs to form our test set (180 images per scene). 
For the envmap-based evaluation, in addition to the 233 test envmaps from \citet{gardner2017learning}, we also import 541 envmaps downloaded from PolyHaven~\citep{polyhaven}.
Note that we do not have the envmap-based ground truth for the real objects, and hence we could only provide qualitative comparison for the envmap-based experiment. 
The details of the dataset constructions can be found in Sec.~\ref{sec:appendix:acquisition} and the data can be obtained from \url{https://zhusz.github.io/TRHM-Webpage/}.

\noindent \textbf{Synthetic Data}. We use the eight widely used scenes from the NeRF-Blender dataset~\citep{mildenhall2020nerf} for experimentation.
We mimic the training and testing setting as in the real scenes: we optimize each scene with the OLAT setting as well, with 112 point light locations (spanning the upper hemisphere) and 100 training camera view points (the same as the original dataset), resulting in 11200 training images in total.
In testing, we evaluate with the OLAT setting with 10 unseen view points from the original test set as well as 10 unseen point light directions, resulting in 100 test samples for each scene.
For envmap-based evaluation, we use similar lighting setups as in the captured lightstage scenarios.

\subsection{Baseline Approaches}

We compare with several most representative state-of-the-art approaches to highlight the strengths of our neural model. 

\begin{itemize}
    \item \textbf{IRON}~\citep{zhang2022iron}, a BRDF-based model, was originally proposed to handle the image data captured under the co-located lighting setting. We implemented and generalized the co-located GGX BRDF toward the general setting where the incident direction and the camera view direction needs not be identical.
    \item \textbf{Inverse-Translucent}~\citep{deng2022reconstructing} devised the BSSRDF model with the learnable parameters for modeling translucent materials. We initialized the geometry with the NeuS surface~\citep{wang2021neus} before optimizing.
    \item \textbf{NRTF-m}~\citep{lyu2022neural} learns a pre-computed transfer field on top of a pre-estimated NeuS~\citep{wang2021neus} surface. 
    We followed their \textbf{m}ulti-lighting training stage and fit the model directly to the \textbf{m}ulti-lighting data, and hence we term it as NRTF-\textbf{m} for contrast to the full NRTF model.
    
    \item \textbf{NeRF-tex}~\citep{baatz2022nerf} incorporates 3 additional dimension of the point light incident direction $\omega_l$ as the feed-in of the light data when compared to NeRF~\citep{mildenhall2020nerf} and could be treated as the volume rendering version of NRTF-m~\citep{lyu2022neural}.

    \item \textbf{Neural-Hints}~\citep{zeng2023relighting}  (with \textbf{OSF}~\citep{yu2023learning}) additionally incorporates two more hints, namely the opacity hints and the highlight hints compared to \citep{baatz2022nerf, yu2023learning}, to improve the high-frequency effects (e.g. hard shadows and specular reflection highlights). 
\end{itemize}

All the approaches are trained on the same data as our proposed TRHM model. 
Note that \citet{deng2022reconstructing, lyu2022neural, baatz2022nerf, yu2023learning, zeng2023relighting} are all proposed specific to the point light setting. When applied to the envmap setting, each pixel on the envmap needs to be enumerated and accumulated. As the speed would be rather slow, we down-sampled the test envmap to the resolution of 16 $\times$ 32 so that it could predict within reasonable amount of time. This also reflects one strength of our model - fast relighting rendering with high resolution envmap. For IRON~\citep{zhang2022iron}, we used Mitsuba3~\citep{nimier2019mitsuba} to conduct image-based rendering and speed evaluation.

\begin{figure*}[t]
\begin{center}
\centerline{\includegraphics[width=\textwidth]{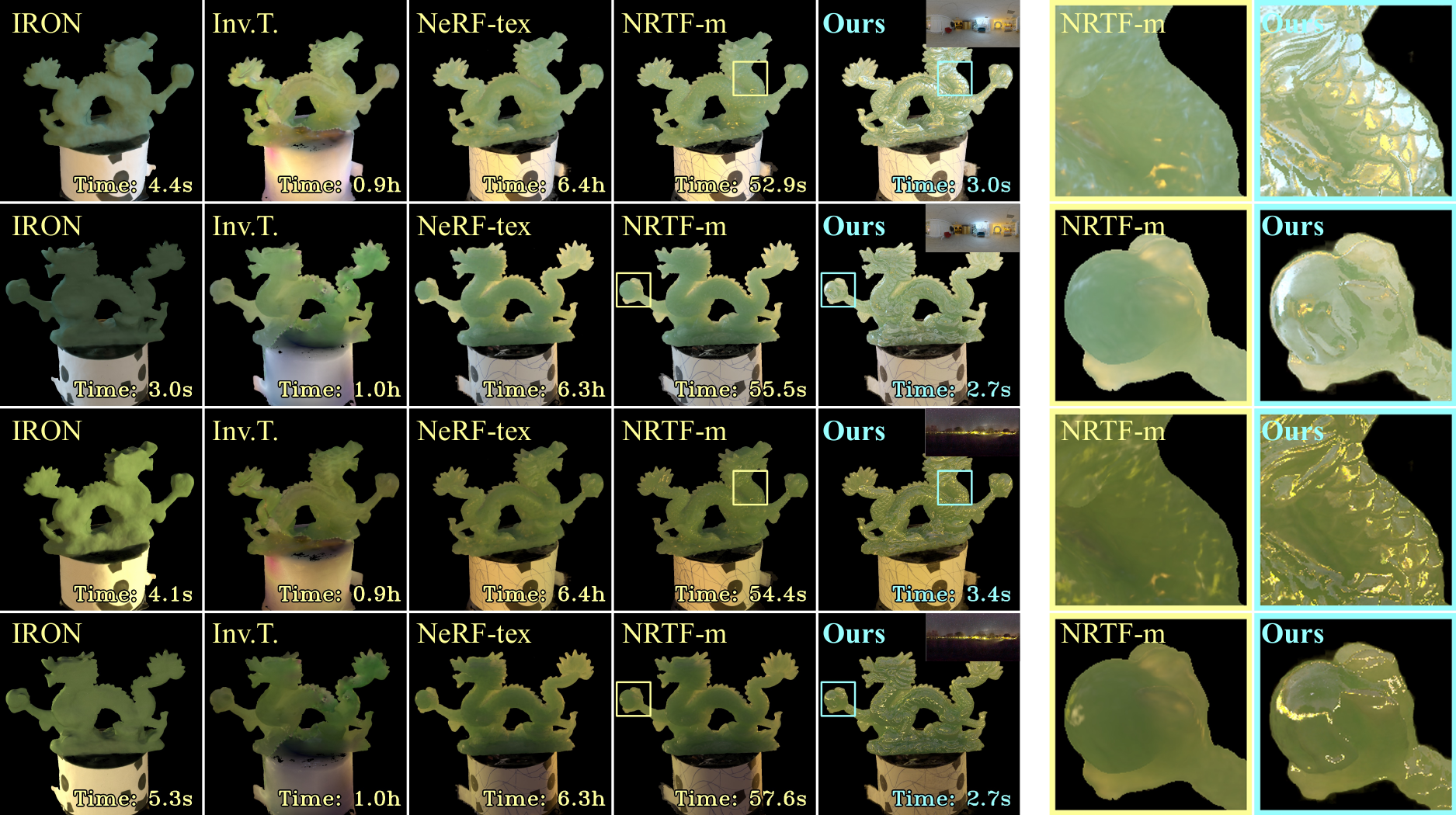}}
\vspace*{-0.3cm}
\caption{Comparing our proposed TRHM model result with representative existing approaches~\citep{lyu2022neural, baatz2022nerf} with respect to the image-based lighting prediction task. While \citet{lyu2022neural, baatz2022nerf} demonstrates the capability of handling subsurface scattering effects, our model nontheless captures local micro geometric details and better demonstrates reflection effects, running at a high speed without the need of enumerate-and-accumulate. Note NeRF-tex~\citep{baatz2022nerf}, NRTF-m~\citep{lyu2022neural} and \textbf{TRHM} (ours) are running at the original full resolution (2048$\times$1366) for qualitative comparison, and all are evaluated with a single Nvidia-3090 GPU.}
\label{fig:qual_exp:envmap}
\end{center}
\end{figure*}

\begin{table*}[t]
\centering
\small
\begin{tabular}{l|ccccc|cc}
\Xhline{4\arrayrulewidth}
 & \multicolumn{5}{c|}{Nvidia 3090}                        & \multicolumn{2}{c}{Nvidia A100}                  \\ \hline
& IRON & Inv-T & NeRF-tex & NRTF-m & \multirow{1}{*}{\textbf{TRHM}} & NRTF-m & \multirow{1}{*}{\textbf{TRHM}}    \\ \hline
Runtime per Frame  & 4.3s & 1.1h & 2.8h & 27s & \textbf{0.72s} & 15s & \textbf{0.30s} \\
\Xhline{4\arrayrulewidth}
\end{tabular}
\vspace{-0.01in}
\caption{Average runtime evaluation (s=seconds, h=hours, $\downarrow$) of our unoptimized implementation over the 800 $\times$ 800 resolution test set with a single Nvidia GPU. Note that our \textbf{TRHM} model can run with less batches of rays on A100 due to the large memory, and hence our approach gets higher acceleration rate on A100 compared to 3090. NRTF-m forwards on the fore-ground pixel only and can forward the whole frame in one batch even on the Nvidia 3090.
}
\label{tab:exp:runtime}
\end{table*}

\subsection{Results}

\noindent \textbf{Qualitative Results}. We provide qualitative comparison in Fig.~\ref{fig:qual_exp:envmap}.
We could see that our results (\textbf{TRHM}) demonstrated clear advantage on lighting effects like subsurface scattering and reflections. 
Note that we achieved such image-based relighting results from only the OLAT image capture data, thanks to the flexibility of our high-frequency hints.
Our modeling of the local micro-geometry also demonstrated clear advantages compared to all the baselines.
In contrast, the enumerate-and-aggregate-based baselines~\citep{deng2022reconstructing, baatz2022nerf, yu2023learning, zeng2023relighting} demonstrates blurred or unclear reflection effects, as the aggregation process tends to average out the sparse and less controllable highlights. 
The BRDF-based baseline approach (IRON~\citep{zhang2022iron}), on the other hand, demonstrates no subsurface scattering effects. Our approach achieves such effects with fast speed, thanks to our hyper-net for enabling subsurface scattering without excessive runtime cost.
Please refer to the appendices for more results.

\noindent \textbf{Runtime}. Our proposed TRHM algorithm demonstrates significant advantages w.r.t. speed as we bypass the necessity of enumerate-and-aggregate over the envmap pixels. We provided the average runtime of our approach as well as all the baselines in Tab.~\ref{tab:exp:runtime}. Our TRHM algorithm is the only candidate that demonstrates sub-second performance on a single Nvidia 3090 GPU for 800$\times$800-resolution rendering, while running at several factors of magnitude faster than all the baselines without the BRDF assumption of the materials (\citet{deng2022reconstructing, lyu2022neural, baatz2022nerf, yu2023learning, zeng2023relighting}),  showcasing its potential for real applications.

\section{Conclusion}

We presented a novel volume-rendering based neural relighting approach adept at handling subsurface scattering effects, strong reflections and translucencies.
Thanks to the proposed modeling of local micro geometries and end-to-end optimization of the radiance transfer gradient on images recorded under various lighting conditions in a light stage, the optimized geometry and appearance reach high quality---even on real data with major subsurface scattering effects.
Furthermore, using a hypernet-based light encoding we enabled fast rendering also in the case of using a general environment map, where existing methods are significantly slower to render.

\bibliography{main}
\bibliographystyle{tmlr}

\appendix

\newpage
\section{Quantitative Experiments}

\begin{figure*}[!b]
\begin{center}
\centerline{\includegraphics[width=0.8\textwidth]{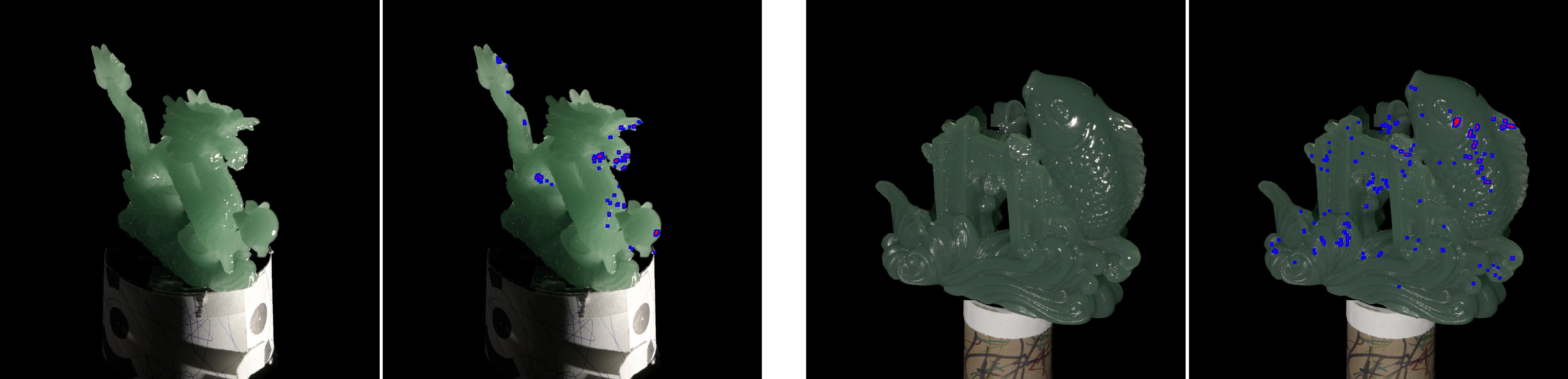}}
\caption{Illustration of the delineation of the highlight pixels and their surrounding pixels (sub-area pixels) on two test cases. In particular, pixels marked as the red color are highlight pixels, while blue for surrounding pixels (sub-area pixels).}
\label{fig:supp:additional_eval_illustration}
\end{center}
\end{figure*}

\begin{figure*}[t]
\begin{center}
\centerline{\includegraphics[width=\textwidth]{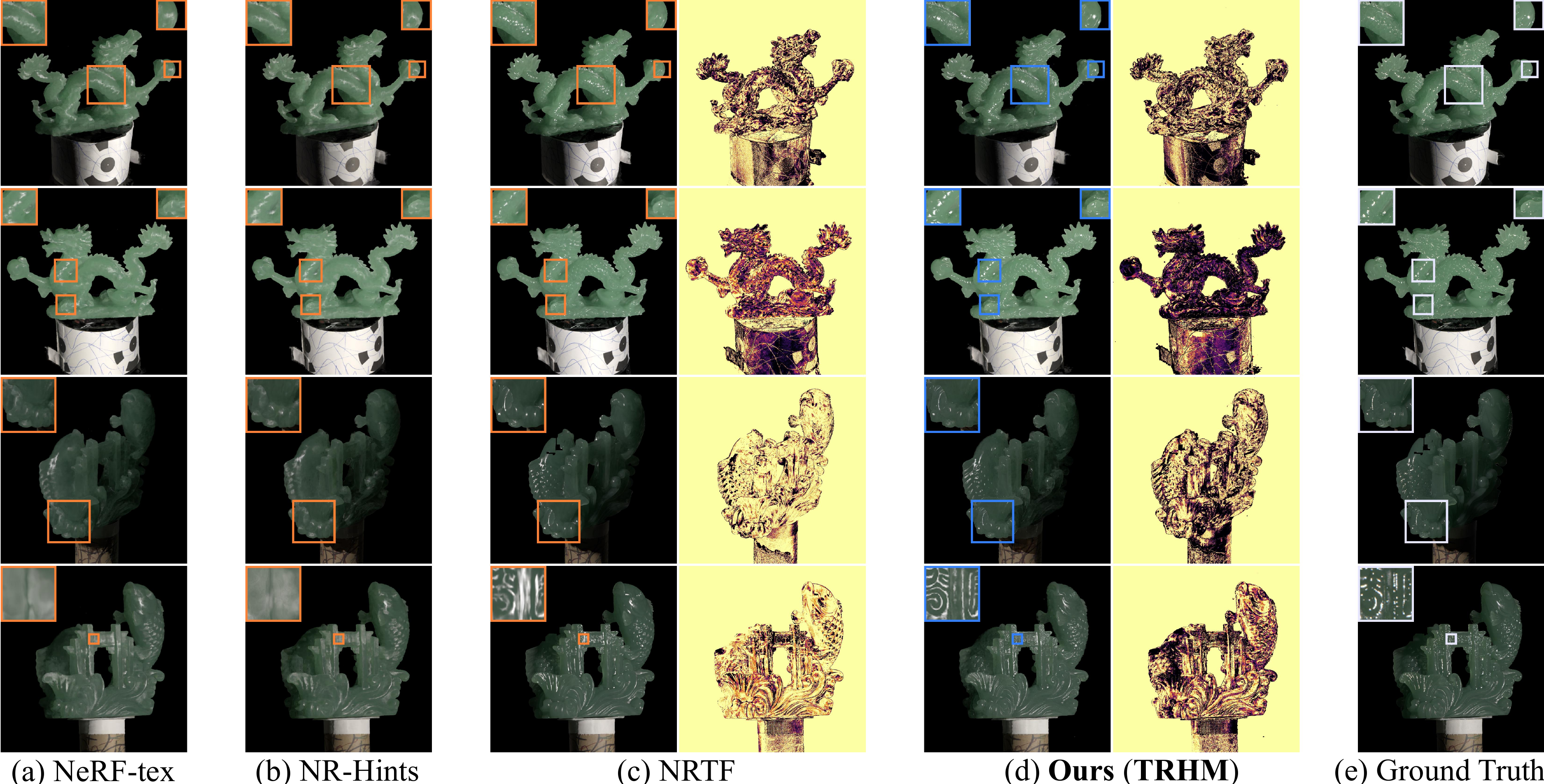}}
\caption{A side-by-side comparison between our proposed \textbf{TRHM} modeling with several most competitive baselines, including NeRF-tex~\citep{baatz2022nerf}, NR-Hints~\citep{zeng2023relighting} and NRTF~\citep{lyu2022neural} under the OLAT evaluation setting. For NRTF~\citep{lyu2022neural} and our \textbf{TRHM}, we further showcase the heatmap of the pixel-wise PSNR (the darker, the worse / lower the PSNR). While for all the visualized four test cases that our PSNR is consistently lower than NRTF as our model is optimized with various regularization terms (Eq.~\ref{eq:lossdotproduct},\ref{eq:losshighfreqreg}) and is not mainly designed for the OLAT relighting task, our visual quality is nonetheless more appealing, especially for specular highlights and local micro geometric modeling. Best viewed electronically with zoom-in. }
\label{fig:supp:additional_quantative}
\end{center}
\vspace{-0.2in}
\end{figure*}

We found conducting quantitative evaluation of the predicted rendering under the envmap lighting condition is challenging, because all the models are trained (optimized) given only the olat training condition before converting to the envmap condition for relighting, causing ambiguity on lighting intensity scaling. 
Furthermore, relighting ground truth for real scene cases under envmap is not available. 
To showcase the performance of our proposed lighting model quantitatively, we instead evaluate the rendering quality in two aspects under the olat condition.
First, we stick to the traditional evaluation setting that evaluate all the pixels in the prediction~\citep{lyu2022neural, mildenhall2020nerf}.
Second, since one of the primary goals of this work is to address the modeling of the lighting effects over glossy surfaces with the particular interest of evaluating specular highlights, we further provide quantitative evaluation of merely the highlight and its surrounding pixels with respect to their accuracy.
More precisely, the criteria of being a highlight pixel is that the HDR intensity is higher than a threshold.
The threshold is scene specific so as to best readjust according to the different lightness between different cases, and the threshold is the same for defining highlight pixels used during training (for Eq.~\ref{eq:lossdotproduct}).
We evaluate highlight pixels as such in particular to showcase the effectiveness of various approaches modeling specular highlights and geometric details.
We further include the surrounding pixels of the highlight pixels, named as sub-area pixels, by dilating the highlight masks, to further test the pixel intensity accuracy surrounding the highlight areas.
A visualization of highlight pixels and its surrounding area is shown in Fig.~\ref{fig:supp:additional_eval_illustration}, and the full details of the delineation is provided with the code release.

\noindent \textbf{Benchmarking and Metrics}. 
Our evaluation metrics follow the standard metric protocol~\citep{lyu2022neural, mildenhall2020nerf}, including PSNR, SSIM and LPIPS~\citep{zhang2018unreasonable} when evaluating the full pixel map. 
When evaluating the highlight and surrounding pixels, only PSNR is reported as these sparse sets of pixels do not constitute a full semantic image.

\subsection{Comparing with existing approaches}

We provide quantitative comparison in Tab.~\ref{tab:exp:ref_olat} and Tab.~\ref{tab:exp:highlight_quant}.
It is worth pointing out that our approach is primarily designed for image-based relighting optimized from the OLAT training data, and hence our approach is not particularly optimized for this evaluation. 
Our approach still obtains competitive performance compared to existing data-driven approaches~\citep{lyu2022neural, baatz2022nerf, zeng2023relighting} when evaluating over all the pixels (Tab.~\ref{tab:exp:ref_olat}), even if our model optimization undergoes several regularization terms (e.g. Eq.~\ref{eq:lossdotproduct}, \ref{eq:losshighfreqreg}).
Our visualization (Fig.~\ref{fig:supp:additional_quantative}) further shows that while our results are not quantitatively higher than NRTF-m~\citep{lyu2022neural}, it nonetheless indicates that our results prevail in the overall visual quality and specular highlight consistency.
On the other hand, our results on highlight pixels evaluation (Tab.~\ref{tab:exp:highlight_quant}) demonstrate advantages compared to several most competitive baseline approaches, showcasing the faithfulness of our modeling specular reflection and local micro geometric details.

\begin{table*}[t]
\centering
\vspace{-0.1in}
\caption{OLAT point-light-based evaluation (PSNR $\uparrow$, SSIM $\uparrow$, LPIPS $\downarrow$) of our approach and the comparison with IRON (BRDF-based)~\citep{zhang2022iron}, Inv-Translucent (BSSRDF-based)~\citep{deng2022reconstructing}, NRTF (Fixed neural surface)~\citep{lyu2022neural}, NeRF-tex~\citep{baatz2022nerf, yu2023learning} and NR-Hints~\citep{zeng2023relighting}. }
\resizebox{1\textwidth}{!}{\begin{tabular}{cccccccc}
\Xhline{4\arrayrulewidth}
     &    & \citep{zhang2022iron} & \citep{deng2022reconstructing} & \citep{lyu2022neural} & \citep{baatz2022nerf} & \citep{zeng2023relighting} & \textbf{TRHM} \\ \hline
\multirow{3}{*}{Dragon} & PSNR  & 17.5 & 21.6  & 31.3 & 30.5     & 30.2     & 30.1 \\
                        & SSIM  & 85.5 & 85.7  & 96.5 & 95.4     & 95.2     & 94.8 \\
                        & LPIPS & .131 & .143  & .068 & .080     & .080     & .079 \\ \hline
\multirow{3}{*}{Fish}   & PSNR  & 20.7 & 23.6  & 33.0 & 31.2     & 31.1     & 31.0 \\
                        & SSIM  & 82.8 & 89.6  & 95.9 & 94.2     & 94.2     & 93.6 \\
                        & LPIPS & .173 & .165  & .090 & .116     & .113     & .109 \\ \Xhline{3\arrayrulewidth} 
\multirow{3}{*}{Chair}   & PSNR  & 26.4	 & 24.1	& 33.9	& 31.9	& 34.1	& 34.0 \\
                        & SSIM  & 93.0	& 87.4	& 97.9& 	95.6&	96.4&	95.8 \\
                        & LPIPS & .081	&.128	&.031	&.067	&.061	&.076 \\ \hline
\multirow{3}{*}{Hotdog}   & PSNR  & 22.9	&24.4&	37.3	&34.7&	36.2	&35.4\\
                        & SSIM  & 91.2	&86.3&	98.3	&96.6&	97.0	&96.5 \\
                        & LPIPS & .103&	.172	&.031	&.065&	.063	&.075 \\ \hline
\multirow{3}{*}{Materials}   & PSNR  & 22.3	&23.2&	35.5	&35.5	&34.9	&36.7\\
                        & SSIM  & 83.5	&82.7	&95.9	&95.9&	95.2	&95.9 \\
                        & LPIPS & .182&	.175&	.074&	.074	&.085	&.077 \\ \hline
\Xhline{4\arrayrulewidth}
\end{tabular}}
\label{tab:exp:ref_olat}
\end{table*}

\begin{table*}[t]
\centering
\caption{OLAT point-light-based evaluation (PSNR $\uparrow$) of our approach and the comparison with NRTF (Fixed neural surface)~\citep{lyu2022neural}, NeRF-tex~\citep{baatz2022nerf, yu2023learning} and NR-Hints~\citep{zeng2023relighting} for highlight pixels and their surrounding pixels (sub-area pixels). We evaluate the PSNR with both the tonemapped LDR scale as well as the original HDR scale (clipped between 0 and 1).}
\resizebox{0.9\textwidth}{!}{\begin{tabular}{clcccc}
\Xhline{4\arrayrulewidth}
                           & PSNR      & \citep{lyu2022neural} & \citep{baatz2022nerf} & \citep{zeng2023relighting} & \textbf{TRHM} \\ \hline
\multirow{4}{*}{Dragon}    & Highlight-pix. (HDR)  & 4.5 & 3.7 & 3.6 & \textbf{6.5}\\
                           & + Sub-pix. (HDR)  & 13.6 & 13.2 & 13.3 & \textbf{13.7}   \\
                           & Highlight-pix. (LDR) & 6.9 & 6.1 & 6.1 & \textbf{8.7} \\ 
                           & + Sub-pix. (LDR)  & 16.6 & 16.1 & 16.1 & \textbf{17.0}\\ \hline
\multirow{4}{*}{Fish}    & Highlight-pix. (HDR)  & 4.5 & 2.4 & 2.3 & \textbf{5.0}\\
                           & + Sub-pix. (HDR)  & \textbf{12.9} & 12.4  & 12.4 & 12.8   \\
                           & Highlight-pix. (LDR) &  6.9 & 4.7 & 4.6 & \textbf{7.4}\\ 
                           & + Sub-pix. (LDR)  & \textbf{15.1} & 14.0 & 14.0 & 14.8 \\
                           \Xhline{4\arrayrulewidth}
\end{tabular}}
\label{tab:exp:highlight_quant}
\end{table*}

\begin{table}[t]
\centering
\caption{Evaluation (PSNR $\uparrow$, SSIM $\uparrow$, LPIPS $\downarrow$) of our approach and the comparison with the five ablation baselines (Sec.~\ref{sec:exp:ablation}). For real objects (dragon and fish) we evaluate the OLAT setting (no ground truth on envmap). For synthetic objects we evaluate the envmap setting. }
\resizebox{0.5\textwidth}{!}{\begin{tabular}{ccccccc}
\Xhline{4\arrayrulewidth}
                        &       & I    & II   & III  & IV            & \textbf{TRHM} \\ \hline
\multirow{3}{*}{Dragon} & PSNR  & 29.6 & 29.5 & 29.3 & 29.3          & \textbf{30.1} \\
                        & SSIM  & 94.5 & 94.3 & 94.4 & 94.3          & \textbf{94.8} \\
                        & LPIPS & .082 & .083 & .078 & \textbf{.077} & .079          \\ \hline
\multirow{3}{*}{Fish}   & PSNR  & 30.7 & 30.5 & 30.8 & 30.9          & \textbf{31.0} \\
                        & SSIM  & 93.3 & 93.0 & 93.4 & 93.5          & \textbf{93.6} \\
                        & LPIPS & .113 & .115 & .111 & .110          & \textbf{.109} \\ 
\Xhline{4\arrayrulewidth}
\end{tabular}}
\label{tab:exp:ablation}
\end{table}

\vspace{-0.1in}
\subsection{Ablation Studies}
\label{sec:exp:ablation}

Several key components in our proposed approach provide essential support for the high quality capturing of the high-frequency lighting effects. 

\textbf{Effects of the normal of the reflection purpose $\mathbf{n}''$.} Our approach utilized an additionally predicted normal for the reflection purpose to better handle local micro geometry of the captured scenes. We provide our ablation study for further showcasing the effectiveness of each component by utilizing the refnerf-predicted normal ($\textbf{n}'$, Ablation I) or the analytical denstiy gradient normal ($\textbf{n}$, Ablation II) for predicting the surface reflection.

\textbf{Effects of the multi-level roughness feed of the envmap.} Our approach devised a multi-level roughness hint for feeding in the envmap for reflection prediction. To illustrate its importance, we experiment with feeding in only the original envmap (the least rough pyramid, Ablation III) or the roughest pyramid (Ablation IV) for predicting the surface reflection.

We showcase the comparison results in Tab.~\ref{tab:exp:ablation}. The results validate the effectiveness of the use of the reflection-purpose normal as well as the necessity of utilizing multiple roughness levels for the envmap-based reflection effects.

\section{More Qualitative Results}

\begin{figure*}[t]
\begin{center}
\centerline{\includegraphics[width=\textwidth]{fig/figure_synthetic_new_3.pdf}}
\vspace*{-0.3cm}
\caption{Further results of our proposed TRHM on synthetic data (reflective cases) under varying view points and image-based lighting.}
\label{fig:supp:synthetica_reflective}
\end{center}
\vspace{-0.2in}
\end{figure*}

\begin{figure*}[!h]
\begin{center}
\centerline{\includegraphics[width=\textwidth]{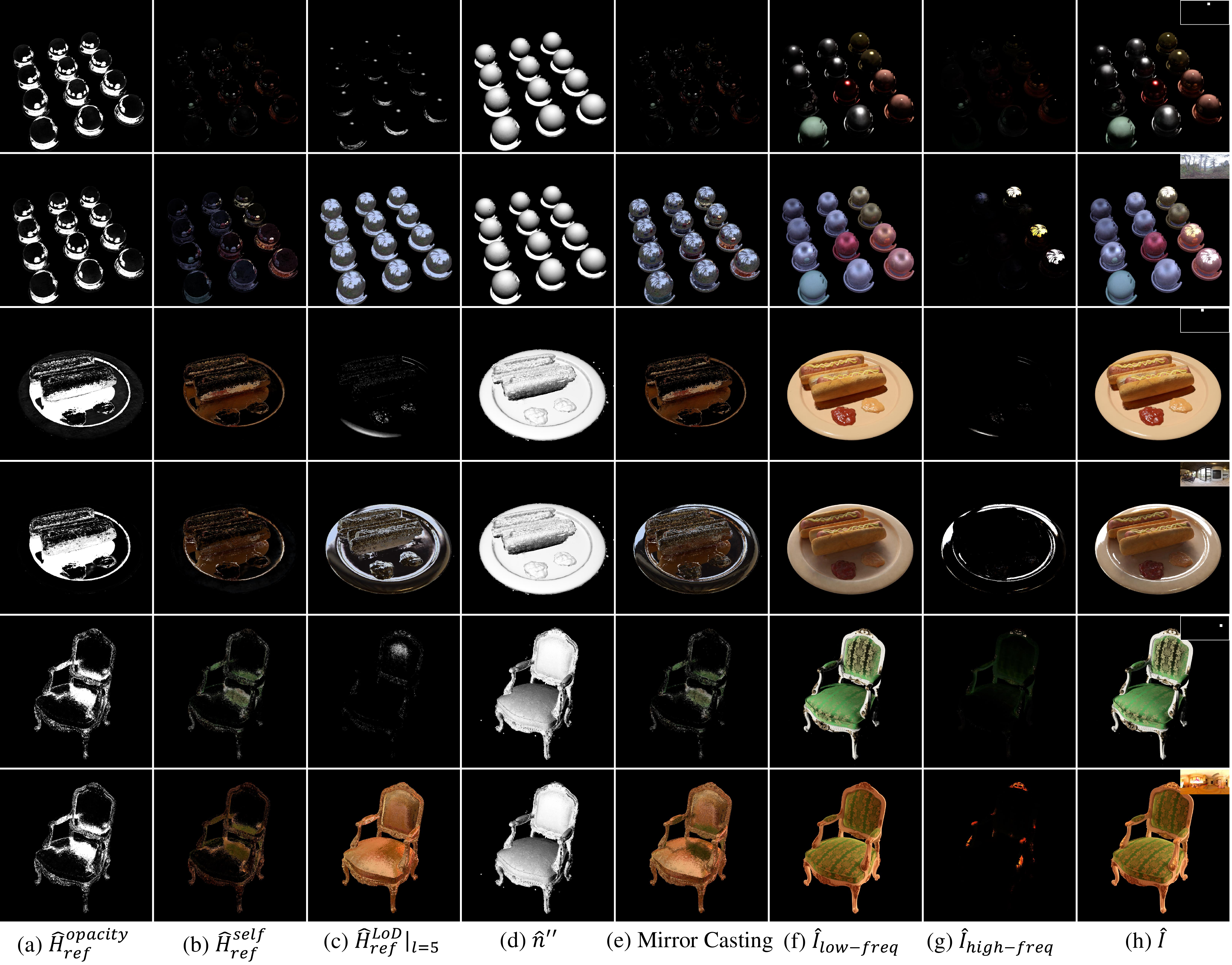}}
\vspace*{-0.3cm}
\caption{Visualization of several important components in our prediction framework. The mirror casting (e) is the visualization that we cast all the materials as the mirror, and reflect the \textit{original} scene appearance as well as the environment map. It is composited by (b) $\hat{H}^\text{self}_\text{ref}$ and (c) $\hat{H}^\text{LoD}_\text{ref}|_{l=0}$, and is switched by (a)$\hat{H}^\text{opacity}_\text{ref}$. It is worth pointing out that the chair scene (Row 5-6) only has a small fraction of the pixels representing reflective materials, and its reflection normal $\hat{n}'$ (d) for non-reflective area is optimized only via tying to $\hat{n}'.$}
\label{fig:supp:synthetic_low_high}
\end{center}
\vspace{-0.3in}
\end{figure*}

\begin{figure*}[t]
\begin{center}
\centerline{\includegraphics[width=\textwidth]{fig/figure_real_diffuse_full.pdf}}
\vspace*{-0.3cm}
\caption{Further results of our proposed TRHM on the non-reflective cases of the light stage data, running at comparable speed with Instant-NGP~\citep{muller2022instant}. Since the computation of the high-frequency branch is unnecessary in this scenario, the time complexity of our approach is the same as Instant-NGP~\citep{muller2022instant}, and the hyper-net forwarding overhead shared by all the pixels is negligible.}
\label{fig:supp:real}
\end{center}
\end{figure*}

\begin{figure*}[!h]
\begin{center}
\centerline{\includegraphics[width=\textwidth]{fig/figure_synthetic_new_1.pdf}}
\caption{Further results of our proposed TRHM on synthetic data (non-reflective cases) under varying view points and image-based lighting.}
\label{fig:supp:synthetica_nonreflective}
\end{center}
\end{figure*}

We further provide visual results of our proposed TRHM framework in Fig.~\ref{fig:supp:synthetica_reflective}-\ref{fig:supp:synthetica_nonreflective}.

\noindent \textbf{Results on synthetic reflective scenes}. Figure~\ref{fig:supp:synthetica_reflective} showcases our results on three reflective cases in the Nerf-Blender dataset (synthetic data). We emphasize that all the frames are predicted within second similar to the real data prediction. Note that we mimic the real light stage training setting on the synthetic data, where our model is only learned from a set of OLAT ground truths, and transfer to the image-based relighting setting. We can see that the results add to our findings that our proposed approach could well conduct image-based relighting at the speed of sub-second.

\noindent \textbf{Component visualizations}. We further visualize several important components in our prediction framework in Fig.~\ref{fig:supp:synthetic_low_high}. We could observe our model could faithfully decompose low frequency and high frequency components into the relative branches ((f) and (g)), with each branch's own strategy to transfer from the OLAT setting (Row 1, 3, 5) to the image-based relighting setting (Row 2, 4, 6). 

\noindent \textbf{Results on non-reflective scenes}. We show other cases without major reflective lighting effects in Fig.~\ref{fig:supp:real} (for real data) and Fig.~\ref{fig:supp:synthetica_nonreflective} (for synthetic data) respectively. It is worth pointing out that while our approach could automatically find out no high-frequency components could be decomposed, we can nontheless switch off the high-frequency branch manually, leading to a very fast prediction pipeline. Particularly, in such circumstance, only the low-frequency branch - an Instant-NGP~\citep{muller2022instant} model whose appearance branch weight is predicted by a hypernet - costs the runtime, and since the hyper-net prediction is shared by all the pixels in the same prediction frame, the runtime of non-reflective scenes of our algorithm stay appoximately the same as that in Instant-NGP.

\noindent \textbf{Limitations.} Due to our assumptions of not handling hard shadows and anistropic lighting effects, our results demonstrate several drawbacks that provide insufficient lighting effect modeling. For instance, the hard shadow for the hotdogs (Row 1 of Fig.~\ref{fig:supp:synthetica_reflective} and Row 4 of Fig.~\ref{fig:supp:synthetic_low_high}) are missing in most cases, and our approach also omits all the reflections in the drums scene (Row 1 of Fig.~\ref{fig:supp:synthetica_nonreflective}). 
We also noticed that the materials scene demonstrate more complicated lighting effects such as metal or pure mirrors, where the desired lighting effects are not obtained. In particular, we notice that the mirror area is mostly dark under the OLAT setting, as the circular light spots are fully concentrated on the sphere and spanning less than one pixel, so that the acquired training data (OLAT) used by our approach cannot effectively capture its lighting effect properties.

\section{Light Stage Data Acquisition}
\label{sec:appendix:acquisition}

\label{sec:capture}

To facilitate studies on the light-dependent appearance modeling of objects and scenes under significant subsurface scattering effects, it is critical to acquire real-world objects featuring such effects.
While existing datasets (e.g.,~\citep{deng2022reconstructing}) includes captures of two translucent objects, they are often limited by resolution and fidelity of the acquired images, causing local micro geometry details to not be fully captured.
To reconstruct a relightable model in a data-driven fashion, we aim to have real-world captures with densely sampled camera viewpoints, complete incident light direction coverage and high-resolution images retaining as much detail as possible.
Consequently, we propose a new dataset, named as TOLC (Translucent-OLAT-Lightstage-Capture), consisting of 11 scenes with significant subsurface scattering effects. 
Our captured data demonstrates high fidelity, preserving rich appearance details, and represents a total of 15TB (3000 times larger than the currently highest quality dataset with similar goals to our knowledge,~\citep{deng2022reconstructing}).

\begin{figure*}
\begin{center}
\centerline{\includegraphics[width=0.95\textwidth]{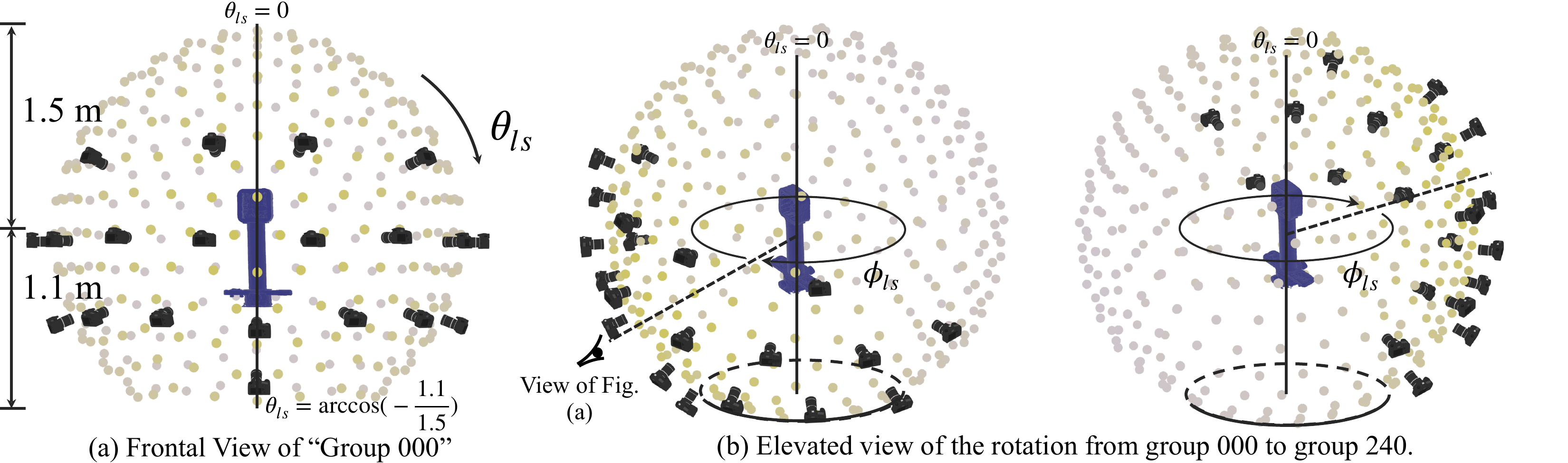}}
\vspace*{-0.4cm}
\caption{Illustration of our light stage capture system. A full capture consists of 9 capture groups, with each group labelled as ``000'', ``040'', ``080'', ..., ``320'', with their number denoting the 40 degree-stepped yaw rotation (see ``$\phi_\text{ls}$'' in (b)). Lights are visualized as dots and cameras with camera icons. All lights are of the same white color---the visualized dot colors merely refer to the light bulb instance, highlighting that the lights are locked with the cameras when rotating between groups. \textbf{(a)}~Frontal view of the system (group ``000''). The radius of the light stage is 1.5 meters, with its center at 1.1m height---the layout is a bottom-truncated sphere. \textbf{(b)}~Rotating from group ``000''~(b-left) to group ``240''~(b-right) according to the ``$\phi_\text{ls}$'' rotation. (b-left) and (b-right) are visualized at an elevated angle. (a) is viewed from the dashed line direction in (b-left).}
\label{fig:lightstage}
\end{center}
\vspace{-0.2in}
\end{figure*}

As shown in Fig.~\ref{fig:lightstage}, we place the cameras and the light sources on the spherical light stage cage, while the objects to be captured are placed on a holding table in the center with a height of roughly 1.1 meter. 
In particular, when capturing the data, our cameras and the light bulbs are fixed on the sphere, while a turntable in the middle can be freely rotated.
Ignoring background pixels, this is equivalent to keeping the object scene static to satisfy the consistency of the scene among views, while rotating the cameras and the light bulbs altogether. 
Throughout the text, we assume that the light stage is configured in the latter case for notational convenience.
Our camera/light-bulb sphere radius is roughly 1.5 meter from the surface of the holding table in the middle).\footnote{Our light bulbs only span roughly between [0, $\frac{3}{2} \pi$) for $\theta_{ls}$, hence no light bulb has a negative altitude even if the sphere radius is larger than the height of the center---the holding table.} 
The rotations of the sphere put the whole captured frames into 9 groups, with each group corresponding to one particular rotated setting of the camera-light sphere. 
On the sphere, we have a total of 20 cameras as well as 331 lighting bulbs (serving as 331 OLAT point lights).\footnote{Notably, since the point light locations are locked with the camera during rotation, the OLAT location in different groups are different from each other. In other words, in our whole dataset, there are only up to 20 images that have been recorded with the same lighting.} 
Consequently, in each group we captured a $20 \times 331 = 6620$ frames, and for the total 9 groups, we captured a total of 59580 frames for one scene.
Our camera captures high dynamic range value for the RGBs, with the cutoff threshold at $4.4019$. 
 The original captured frames come with a resolution of $8192 \times 5464$. We found a 4 times down-scaling retains most of the texture details and hence we conduct all our experiments on the down-scaled version ($2048 \times 1366$). Notably, all the captures at the resolution of $2048 \times 1366$ still span 15TB of storage. 
During the capture, the cameras always face toward the objects on the holding table, and we tune the focal length of the camera to best suit the size of the particular objects. 
We obtain the extrinsic camera poses via an off-the-shelf software with manual corrections. 
Since the light bulbs shining in the opposite direction of the camera incur significant noise to the reconstruction process (especially considering that the rotation between the group would make the background inconsistent), we introduced several heuristics, including RGB variations and saturation to segment out the background.  
All the camera poses, light locations as well as the masking information are used by all the approaches in our evaluation sections, and we shall make all the details about the data publicly available to facilitate future research.

\end{document}

%% file: math_commands.tex
\usepackage{amsmath,amsfonts,bm}

\def\eqref#1{equation~\ref{#1}}

\def\1{\bm{1}}

\DeclareMathAlphabet{\mathsfit}{\encodingdefault}{\sfdefault}{m}{sl}
\SetMathAlphabet{\mathsfit}{bold}{\encodingdefault}{\sfdefault}{bx}{n}

